\documentclass[letterpaper, 10 pt, conference]{ieeeconf}  

\IEEEoverridecommandlockouts                              
\usepackage[hypcap=true]{subcaption}
\usepackage[dvipsnames]{xcolor}

\definecolor{flatgreen}{HTML}{2ECC71}
\definecolor{flatpurple}{HTML}{9B59B6}
\definecolor{deepskyblue}{HTML}{00bfff}

\renewcommand{\baselinestretch}{0.98}

\usepackage{amsmath}
\usepackage{amssymb}
\usepackage{amsfonts}
\usepackage{amsbsy}
\usepackage{mathrsfs}

\DeclareMathOperator*{\argmin}{arg\,min}
\usepackage{mathtools}

\usepackage{float}

\usepackage{url}

\newcommand{\methodname}{THOR + G-ICP}

\newcommand{\NNN}{{\mathbb{N}}}

\newcommand{\RRR}{{\mathbb{R}}}

\usepackage[colorinlistoftodos]{todonotes}
\let\xtodo\todo
\renewcommand{\todo}[1]{\xtodo[inline,color=black!5]{#1}}

\title{\LARGE \bf
Deep 6-DoF Tracking of Unknown Objects for Reactive Grasping
}

\author{Marc Tuscher$^{1, 2}$ \and Julian Hörz$^{2}$ \and Danny Driess$^{2, 3}$ \and Marc Toussaint$^{3, 4}$
\thanks{$^{1}$ sereact.}
\thanks{$^{2}$ Machine Learning and Robotics Lab, University of Stuttgart.}
\thanks{$^{3}$ Max-Planck Institute for Intelligent Systems, Stuttgart.}
\thanks{$^{4}$ Learning and Intelligent Systems, TU Berlin.}
}

\begin{document}

\maketitle
\thispagestyle{empty}
\pagestyle{empty}

\begin{abstract}
Robotic manipulation of unknown objects is an important field of research.
Practical applications occur in many real-world settings where robots need to interact with an unknown environment.
We tackle the problem of reactive grasping by proposing a method for unknown object tracking, grasp point sampling and dynamic trajectory planning. 
Our object tracking method combines Siamese Networks with an Iterative Closest Point approach for pointcloud registration into a method for 6-DoF unknown object tracking.
The method does not require further training and is robust to noise and occlusion.
We propose a robotic manipulation system, which is able to grasp a wide variety of formerly unseen objects and is robust against object perturbations and inferior grasping points.
\end{abstract}

\section{Introduction}\label{sec:intro}
Aiming towards a wide-spread application of robots in natural or unstructured industrial environments, methods for robotic perception need to incorporate more general approaches.
Robots manipulating in noisy and cluttered real-world scenarios require the ability to interact with arbitrary, unknown objects.
Interacting with objects in the real world requires the ability to efficiently perceive the target object.
Target objects need to be tracked in real-time. 
Versatile and robust tracking algorithms are required.
Further, in order to act in a dynamically changing environment, robots need to react to changes and collaborate with other agents.

This work tackles the problem of reactive grasping of unknown objects.
That is, grasping of unknown objects under perturbation and changes of the environment.
We divide the problem of reactive grasping into three subproblems: detection and real-time tracking of unknown objects, computation of grasp configurations and reactive trajectory planning.

\begin{figure}[t]
	\centering
	\begin{subfigure}[c]{.238\textwidth}
		\includegraphics[width=1.\textwidth]{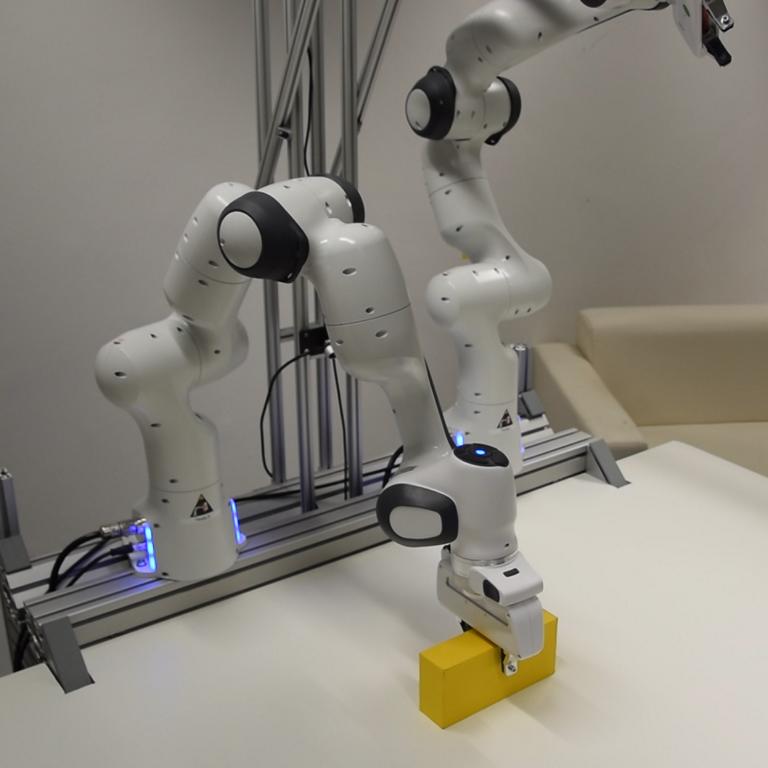}
		\subcaption{Yellow cuboid}
		\vspace{-8pt}
  \end{subfigure}
  \hfill
	\begin{subfigure}[c]{.238\textwidth}
		\includegraphics[width=1.\textwidth]{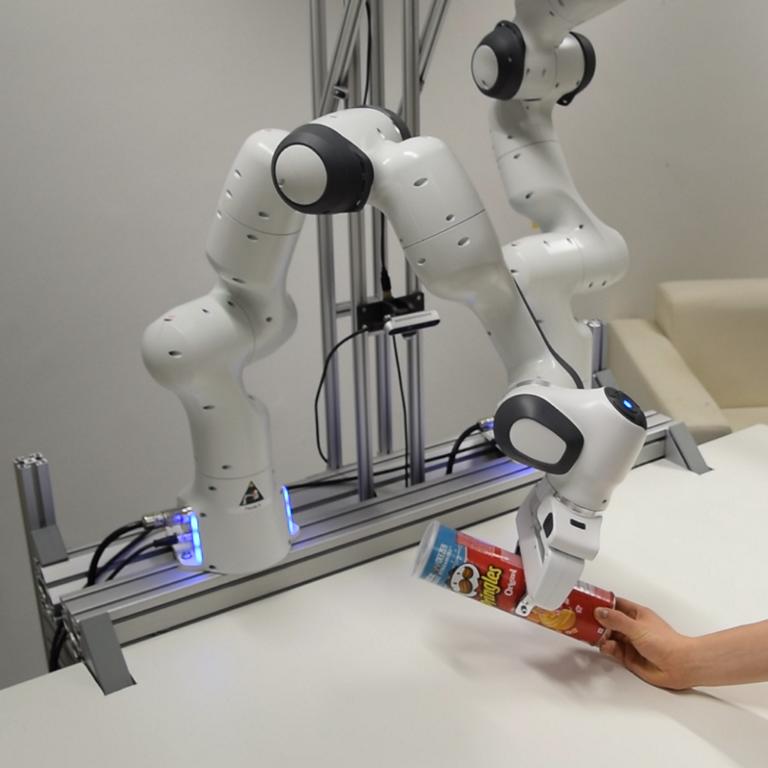}
		\subcaption{Pringles can}
		\vspace{-8pt}
  \end{subfigure}
  \vskip\baselineskip
	\begin{subfigure}[c]{.238\textwidth}
		\includegraphics[width=1.\textwidth]{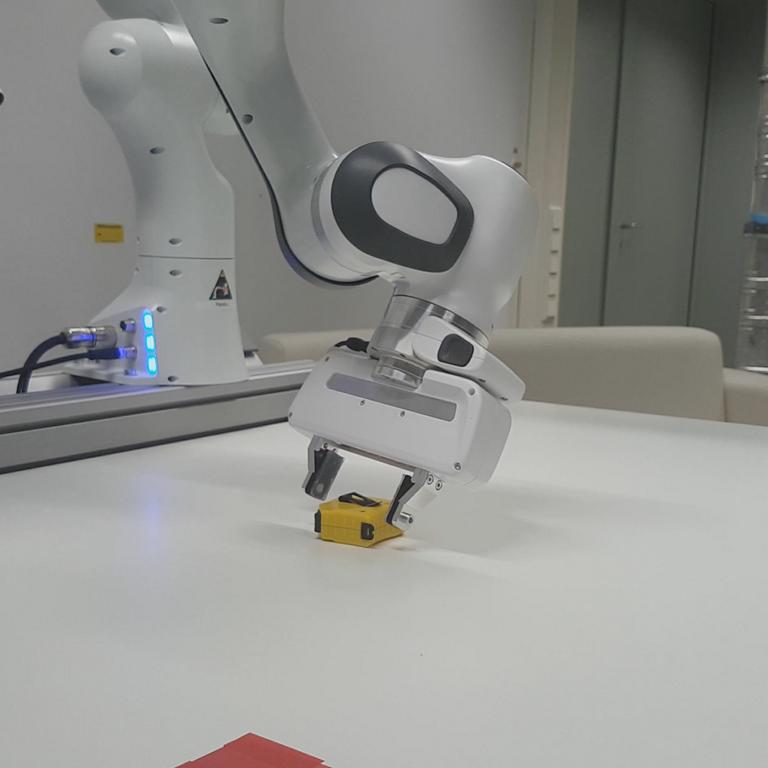}
		\subcaption{Measuring tape}
		\vspace{-2pt}
  \end{subfigure}
  \hfill
	\begin{subfigure}[c]{.238\textwidth}
		\includegraphics[width=1.\textwidth]{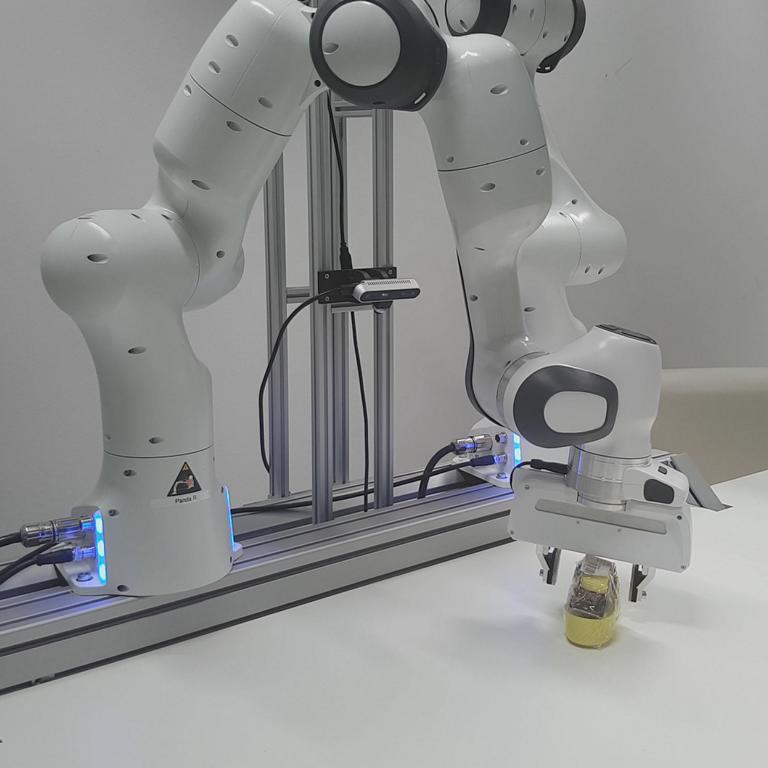}
		\subcaption{Lashing strap}
		\vspace{-2pt}
	\end{subfigure}
	\caption{Tracking and grasping unknown objects.}
	\label{fig:objectstograsp}
	\vspace{-20pt}
\end{figure}
Object tracking methods are often trained on specific classes of objects~\cite{tremblayDeepObjectPose2018,giancolaLeveragingShapeCompletion2019} or require a 3D model of the target object a priori~\cite{wuthrichProbabilisticObjectTracking2013,garonDeep6DOFTracking2017, schmidtDARTDenseArticulated2014,schulmanTrackingDeformableObjects2013}.
This heavily limits their usability for practical applications.
Existing methods for unknown object tracking either target coarse grained tracking~\cite{feldmanMultiICPTrackerOnline,choRealTimeObjectTracking2006,almeidaRealTimeTrackingMoving2005,fleder3DObjectTracking}, are prone to pose drifting~\cite{leebMotionNets6DTracking2019} or limited to a specific environment~\cite{kragicVisionRoboticObject2005}.
Existing methods for grasp point detection are too slow for for real-time evaluation~\cite{mahlerDexNetDeepLearning2017} or do not take the current robot configuration into account~\cite{morrison2018closing, kumra2019antipodal}.
Common methods for trajectory planning are designed for static scenes.
Specifically, standard Rapidly Random Trees for trajectory planning are designed for static scenes only.

We propose a method for 6-DoF tracking of unknown objects based on Siamese Networks for 2D tracking and Iterative Closest Point (ICP) for pointcloud registration.
This method is able to track the 6-DoF pose of a wide variety of objects.
That is, given a set of RGB templates captured during tracking and a pointcloud template of the object captured at an initialization phase, the method outputs the subsequent poses of the target object in real-time.
Further, we propose a method for grasp candidate computation and evaluation, taking the depth of the target object and the current robot state into account.
We also propose a method for reactive trajectory planning in the presence of dynamic changes and collisions.
This system is able to efficiently manipulate a diverse set of objects. 
We demonstrate this ability in our accompanying video\footnote{\label{foot:video}\url{https://youtu.be/Hew00rMw8qg}}.
We summarize our contributions as follows: 
\begin{enumerate} 
\item Our approach combines Siamese Networks for RGB tracking with ICP to realize a conceptually simple, but powerful 6-DoF tracking method: this method is able to track a wide variety of unknown objects under occlusion in real-time, while being class-agnostic and model-free.
\item We propose a novel method for reactive grasping, able to grasp dynamic objects even under perturbation of the robot or the target object.
\item  We combine both approaches with a method for reactive trajectory planning accounting for dynamic changes of the environment.
\end{enumerate}

The paper is structured as follows: in section~\ref{sec:related} related work is presented, section~\ref{sec:background} provides background on the methods we built upon, section~\ref{sec:tracking} explains our object tracking methond and section~\ref{sec:grasping} describes the approach to reactive grasping.
In section~\ref{sec:experiments} experiments are conducted in simulation and on a real robot.
\section{Related Work}\label{sec:related}

\subsection{Object Tracking}

\paragraph{Model-based tracking}
One predominant paradigm for 6-DoF object tracking is model-based object tracking.
That is, a 3 dimensional description of the target object is available a priori.
A large body of work has been evaluated on this topic.
Particle Filtering~\cite{doucetSequentialMonteCarlo} proves to be useful in model-based object tracking~\cite{vermaakMaintainingMultimodalityMixture2003}.
Particle filter based tracking models often operate directly on pointclouds.
Incorporating control inputs from the robot, to give a good estimate on the current object is also successfully applied to particle filter trackers~\cite{wuthrichProbabilisticObjectTracking2013}.
Other appraoches extend particle filter tracking to multi-target tracking using a mixture in particles~\cite{vermaakMaintainingMultimodalityMixture2003} and learning the correct correspondence to a certain target object via AdaBoost~\cite{okumaBoostedParticleFilter2004}.
\cite{rusuradub.Tracking3DObjects2012} uses a particle filter algorithm operating on RGB-pointclouds, thus, integrating texture information into tracking.

Fully integrating the robot kinematics in object tracking helps to reduce the degrees of freedom of the target object and serves as a good prior for vision based object tracking in robotic manipulation~\cite{schmidtDARTDenseArticulated2014,gaoFilterRegRobustEfficient2019}.
However, such approaches rely on both, an object model and a robot model being present beforehand, and reduce the general applicability of the system.

Rendering multiple simulated views from the 3D model of the target object can be used to leverage deep transfer learning for pose estimation~\cite{xiaoPoseShapeDeep2019}.
Garon~et.~al~\cite{garonDeep6DOFTracking2017} use convolutional neural networks to directly operate on RGBD data, by training from scratch for each individual object.
SegICP~\cite{wong2017segicp} successfully combines deep neural networks for semantic segmentation with ICP for known object tracking.

\paragraph{Pose estimation}
Pose estimation and Object Detection are related fields of research: in \cite{erolImprovedDeepNeural2018} a deep learning method for object detection is extended by an object database yielding an object tracking system.
Many approaches extend deep learning models for visual object detection to 6-DoF pose estimation from RGB data~\cite{pavlakos6DoFObjectPose2017,songDeepSlidingShapes2016,hou3DSIS3DSemantic2019}.
Applying deep learning models directly on pointcloud data seems promising since the rise of PointNet feature extractors~\cite{qiPointNetDeepLearning2017,qiPointNetDeepHierarchical2017}.
VoteNet~\cite{qiDeepHoughVoting2019} combines deep learning with Hough voting~\cite{leibeCombinedObjectCategorization} to predict 3D bounding boxes on pointcloud input data.

However, most approaches to pose estimation lack the ability to run in real-time, and therefore render impractical for robotic manipulation.
Tremblay~et.~al~\cite{tremblayDeepObjectPose2018} train a deep neural network for pose estimation on synthetic RGB data using domain randomization.
The method is suited as a real-time system for real-world robotic grasping of known objects.
All approaches in this paragraph, however, are only able to track objects from a set of predefined classes.

\paragraph{Unknown objects}
Despite the practical importance, only little research is done on tracking of unknown objects in robotics applications.
Early work uses optical flow tracking and an integrated eye-in-hand vision system to grasp arbitrary objects with a visual servoing appraoch~\cite{luoAdaptiveRoboticTracking1988}.
Experiments show that the approach only works in a specific environment.
Other approaches rely on detecting and segmenting unkown objects in a known environment~\cite{kragicVisionRoboticObject2005} or do not take the shape of the object into account~\cite{pieropanRobust3DTracking2015}.
Methods based on a large number of different algorithms, each tuned for a specific application, are usually brittle and sensitive to changing environments.
A more recent approach to unknown object tracking applies multiple deep learning models to first segment the scene into model segments, then predict the position and orientation relative to the last frame~\cite{leebMotionNets6DTracking2019}.
Therefore, the method bootstraps the current object pose on earlier estimations.
Experiments show that this approach is particularly prone to tracking drift: stable tracking of real-world objects is only possible for approximately one second.

\subsection{Grasp Candidate Detection}

Deep learning for robotic perception has changed the field of computer vision based grasp candidate detection.
Existing approaches show that using neural networks on RGBD input successfully discriminate grasp candidates to evaluate the best grasp to manipulate objects~\cite{mahler2019learning, driess2020deep}.
Neural nets are also used to generate 6-DoF grasps enabling more informed grasping and manipulation of unknown objects~\cite{mousavian2019graspnet}.
Recent work concentrates on the generation of dense pixel-wise quality maps and pixel-wise grasp information to generate multiple grasps at once~\cite{morrison2018closing}.
These approaches use small neural networks to achieve real-time generation of grasp candidates and repeatedly set a new state-of-the-art on international grasping benchmarks~\cite{kumra2019antipodal}.
However, such methods do not take the robot configuration into account.

\subsection{Dynamic Trajectory Planning}

Reactive planning is formerly shown to perform well in complex manipulation tasks~\cite{kappler2018real}.
Dynamic changes of the environment can also be efficiently handled by sampling based approaches~\cite{otte2016rrtx}.

Our method for object tracking combines Siamese Networks for template-matching in RGB frames with ICP-based pointcloud registration.
Using this object tracking method we disentangle perception from grasp point calculation, providing the flexibilty to exectute complex queries on the scene and taking the current robot configuration into account.
We further implement a variant of the conceptually simple bug algorithm in configuration space able to plan collision-free trajectories in a dynamic environment.
\section{SiamMask \& THOR}\label{sec:background}
The method for 6-DoF tracking proposed in this work is built on SiamMask~\cite{wangFastOnlineObject2018} and THOR~\cite{sauerTrackingHolisticObject2019} for template-matching based RGB object tracking.
This section provides background on these methods.
For more details please consider the original papers.

Given a template $T$ and an input $x$, SiamMask computes a cross-correlated feature map
\begin{equation}\label{eq:theoretical:siammask:inference}
g_{\theta}(T,~x) = f_{\theta}(T) \star f_{\theta}(x),
\end{equation}
where $f_{\theta}$ denotes the feature extraction model used to process both, the novel input and the template.
Equation~\ref{eq:theoretical:siammask:inference} yields a bounding box in image coordinates and a tracking score.
Computing a segmentation mask of the target object in the RGB observation is done using a separate branch in the neural network for mask refinement.

THOR consists of a long-term and a short-term module of RGB templates which can be used alongside SiamMask.
In each iteration of tracking a subregion $x$, corresponding to the estimated position and size of the target object, is cropped from the current frame and fed through the SiamMask model
alongside each individual RGB template from the THOR long-term module (LTM) $T_{l1}, \dots, T_{l5}$ and short-term module (STM) $T_{s1}, \dots, T_{s5}$
\begin{equation}\label{eq:theoretical:thor:inference}
g_{\theta}(T_{l1},~ \dots,~ T_{l5}; T_{s1},~ \dots,~ T_{s5}; x).
\end{equation}
Templates in the STM are updated in a first-in first-out manner.
Every $10$ iterations a new template is generated from the current crop and appended to the STM, while the oldest template in the STM is removed.
$T_{l1}$, the ground truth template, remains in the LTM.

Let $z_i = f_{\theta}(T_i)$ denote the features extracted from template $T_i$.
A Gram Matrix
\begin{gather*}
\begin{bmatrix}
  z_1\star z_1  & z_1 \star z_2 & \cdots & z_1 \star z_m \\
  \vdots        & \vdots        & \ddots & \vdots        \\
  z_m \star z_1 & z_m \star z_2 & \cdots & z_m \star z_m
\end{bmatrix}
\end{gather*}
is constructed, where each entry $G_{ij}$ expresses the similarity of templates $T_i$ and $T_j$ by cross-correlating the features $z_i$ and $z_j$.
If a novel template $T_{c}$ is similar enough to the ground-truth template $T_{l1}$ and replacing an existing template in the LTM increases the volume of the parallelotope
$\Gamma\left(z_1,~\dots~,z_5\right)$
spanned by the feature vectors of each template, it is added to the LTM.
A new template $T_{c}$ is similar enough, if its features $z_c$ satisfy the inequality
\begin{equation}
z_c \star z_1 > l \cdot G_{11} - \gamma,
\end{equation}
where $l$ is a hyperparameter to trade-off tracking performance against drifting robustness.
$\gamma$ is computed using the templates in the STM
\begin{equation}
\gamma = 1 - \frac{2}{N(N+1)G_{st, max}} \sum^{N}_{i<j}{G_{st, ij}}
\end{equation}
where $G_{st}$ is a Gram Matrix for the STM similar to $G$.

\section{6-DoF Tracking}\label{sec:tracking}
This section describes the approach this work  takes on tracking arbitrary objects in 6 degrees of freedom.
Since this approach is model-free, only information from a single RGBD camera at initialization time is used to describe the target object.
Initialization of the tracking method requires an initial (arbitrarily defined) object pose in world coordinates and a 2D bounding box in image coordinates.
We assume that we are able to detect an initial pose and a 2D bounding box of an object.

\begin{figure}[!tb]
\centering
\vspace{10pt}
\includegraphics[width=.45\textwidth]{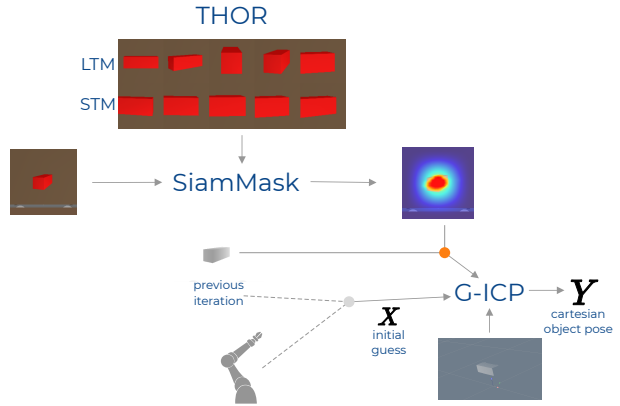}
\caption [Overview of the system.]{\centering Overview of our object tracking component.}
\label{fig:theoretical:system}
\vspace{-15pt}
\end{figure}

Briefly, our method works as follows: \begin{enumerate} \item In an initialization phase, 3D bounding boxes and cartesian poses of arbitrary objects on a table are detected using background subtraction in depth images from a top down view.
\item SiamMask + THOR is started to track the target object in RGB frames. 
An initial pointcloud is saved for 3D pose estimation via pointcloud registration.
\item SiamMask tracks the target object in new frames in real-time, a pointcloud of this observation is segmented from the depth image.
The initial pointcloud is fit to the new observation using Generalized-ICP~\cite{segal2009generalized} to retrieve the 6-DoF transform.\end{enumerate}
Figure~\ref{fig:theoretical:system} shows an overview of the tracking system.
The system maps an RGBD input alongside an initial guess (gained from the previous iteration or a kinematic map) of the target object pose to an estimate of the current object pose.
We call our object tracking method \methodname{}.

\paragraph{Initialization}
RGB tracking is initialized by supplying an initial frame and a bounding box containing the target object to the tracker.
The bounding box is used to crop the frame to an initial RGB template $T_{init}$.
The THOR long-term module (LTM)
\begin{equation}
T_{l1},~ \dots,~ T_{l5}  \leftarrow T_{init}
\end{equation}
and short-term module (STM)
\begin{equation}
T_{s1},~ \dots,~ T_{s5}  \leftarrow T_{init}
\end{equation}
are filled with this initial template of the object.

In order to find a template pointcloud of the target object, the segmentation mask provided by SiamMask in the first iteration of RGB tracking is used to retrieve the pixels $d_{ij} \in \RRR$ in the depth image, which correspond to points on the target object's surface.
We project the depth values $d_{ij}$ to points $x^k \in \RRR^3$ in world coordinates using the inverse camera projection $\hat{P} \in \RRR^{3 \times 4}$ via
\begin{equation}\label{eq:theoretical:pinv}
x^k = \hat{P} \cdot \left(i \cdot d_{ij}~~j \cdot d_{ij}~~d_{ij}~~1 \right)^T.
\end{equation}
The set of these points
$\mathcal{P}_{temp} = \{x^k\}_{k=1}^N$
is the template pointcloud, which is registered to a novel observation to retrieve the transformation of the target object in world coordinates.

\paragraph{Tracking}
RGB tracking is performed according to SiamMask with THOR as template module.
In every step, SiamMask relies on cropping the correct subwindow from the image to perform template matching on a smaller region of the image.
Cropping the correct subwindow, depends on prior information on the target object's location and its size in the image frame.
If the robot is not moving the object, the position and size of the object in the previous frame is used.
If the robot is moving the object, a kinematic map $\phi:~ \RRR^n \rightarrow \RRR^3$ is used to compute an estimate of the current object pose and a 3D bounding box containing the object.
A 2D bounding box in image coordinates is then found by projecting the corners of the 3D bounding box to image coordinates and computing the bounding rectangle containing all corners.
A step of RGB tracking yields a segmentation mask containing the target object in the current frame.
This segmentation mask is used to select the corresponding pixels in the depth image.
These pixels are then projected to points on the object's surface in world coordinates using equation~\ref{eq:theoretical:pinv}.
This set of points $\mathcal{P}_{obs}$ is used as an observation to fit the template pointcloud $\mathcal{P}_{temp}$.

To reduce the computational cost of pointcloud registration, both pointclouds are preprocessed by voxel grid filtering, reducing the dimesionality of the pointcloud while keeping the geometrical properties.

Pointcloud fitting is done using the Generalized-ICP algorithm, setting either the last known pose of the object or the pose computed using a kinematic map, in case the robot is handling the target object, as an initial guess.
Formally, this can be seen as a function mapping from a set of pointclouds and an initial guess $X$ to final transformation $Y$
\begin{equation}
\begin{split}
  h_{\text{G-ICP}}&:~ \RRR^{n \times 3} \times \RRR^{m \times 3} \times \RRR^{4 \times 4} \rightarrow \RRR^{4 \times 4}\\
  h_{\text{G-ICP}}&:~ (\mathcal{P}_{obs}, \mathcal{P}_{temp}, X) \mapsto Y.
\end{split}
\end{equation}
We use this 6-DoF tracking method to perform reactive grasping on a real robot.

\section{Reactive Grasping}\label{sec:grasping}

In this section we develop a method to grasp (cuboid-shaped) objects, which is able to dynamically react to disturbances, both in the environment and the robot configuration.
While our tracking method is able to process arbitrarily defined poses, this method requires the pose to be sensibly defined.
That is, the coordinate system axes should be aligned with the edges of a bounding box surrounding the target object.

\begin{figure*}
	\centering
	\begin{subfigure}[t]{0.23\textwidth}
		\centering
		\includegraphics[width=1.\textwidth]{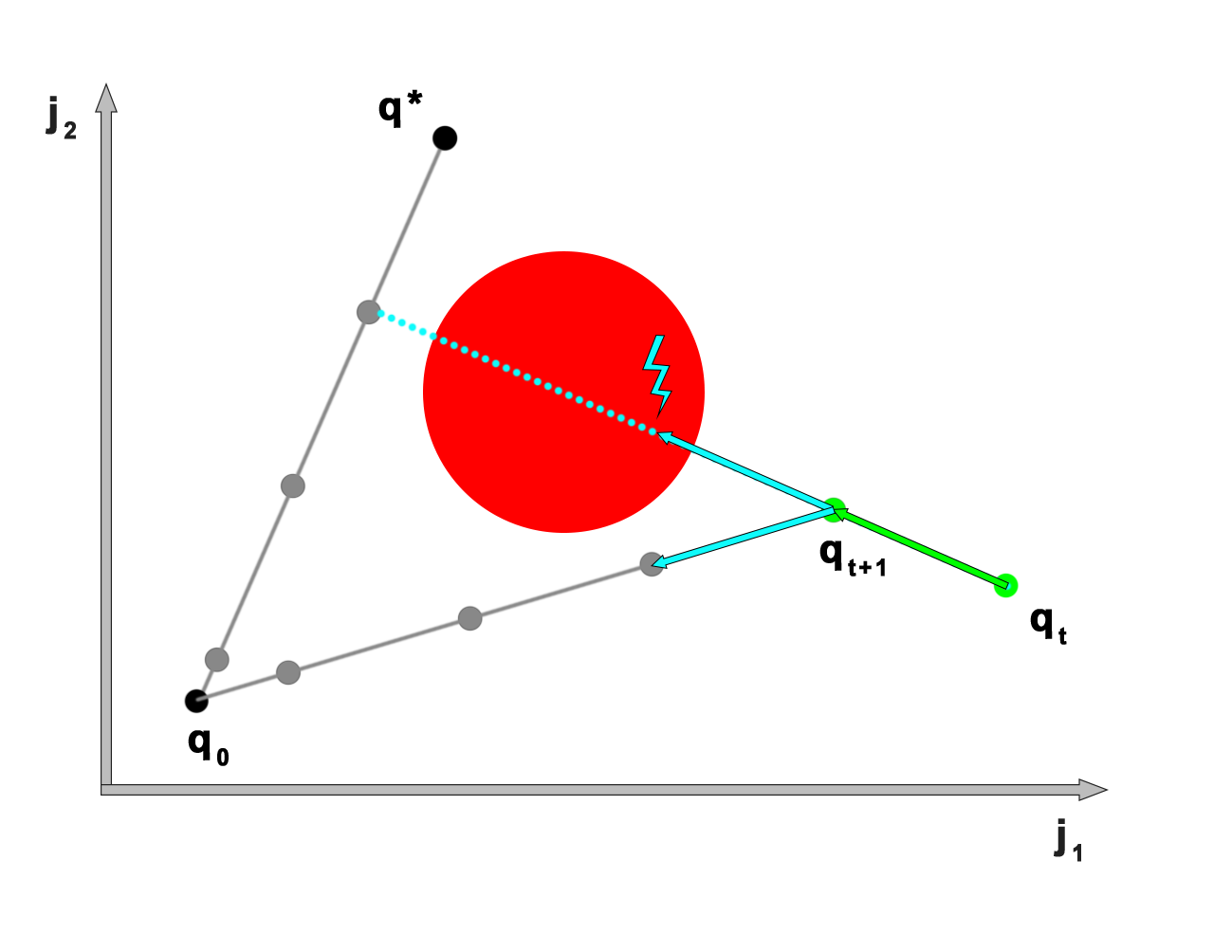}
		\vspace{-25pt}
		\subcaption{Linearly interpolating between $q_t$ and $q^{\ast}$ would lead to collisions with the object.}
	\end{subfigure}
	\hfill
	\begin{subfigure}[t]{0.23\textwidth}
		\centering
		\includegraphics[width=1.\textwidth]{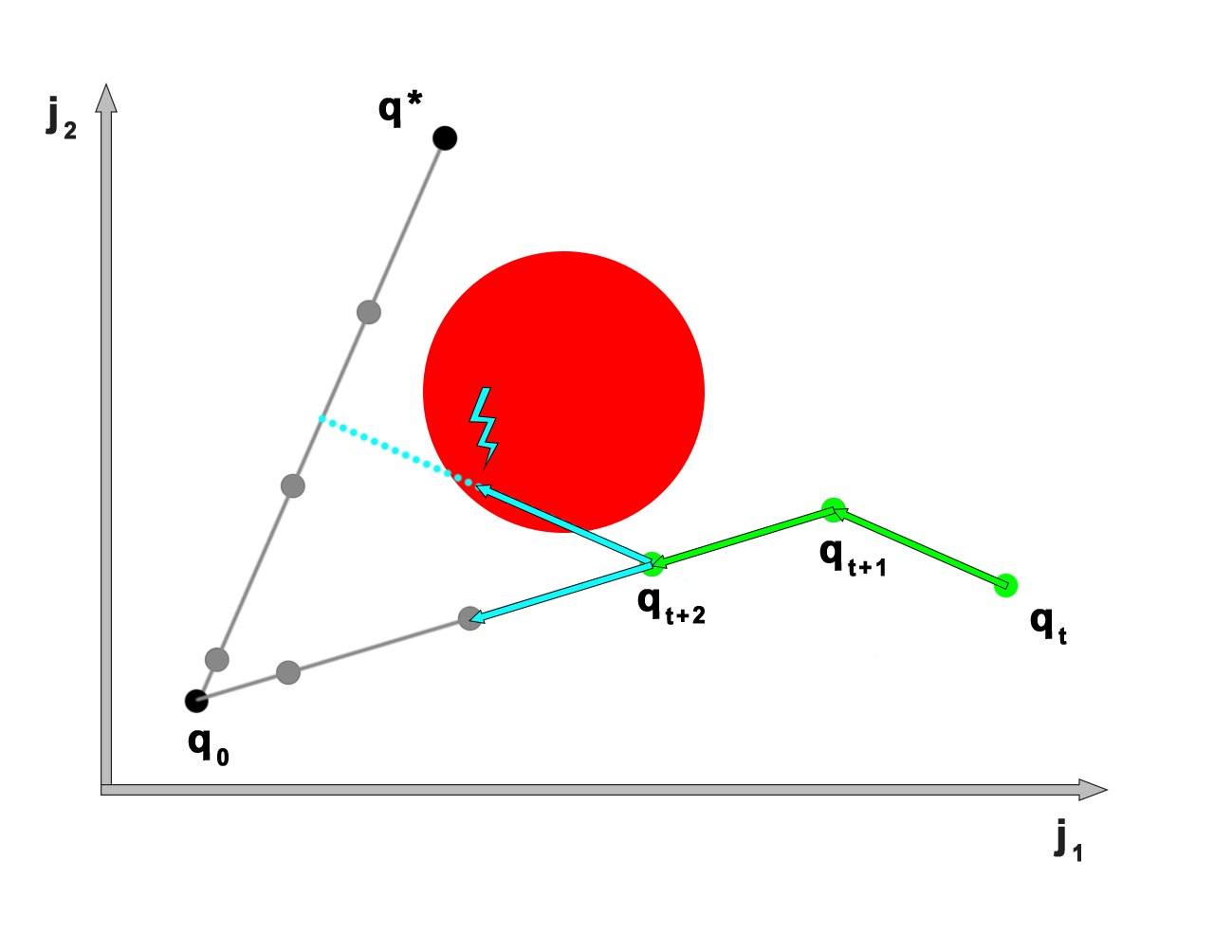}
		\vspace{-25pt}
		\subcaption{After backstepping two steps, the shortest path would still end in collisions.}
	\end{subfigure}
	\hfill
	\begin{subfigure}[t]{0.23\textwidth}
		\centering
		\includegraphics[width=1.\textwidth]{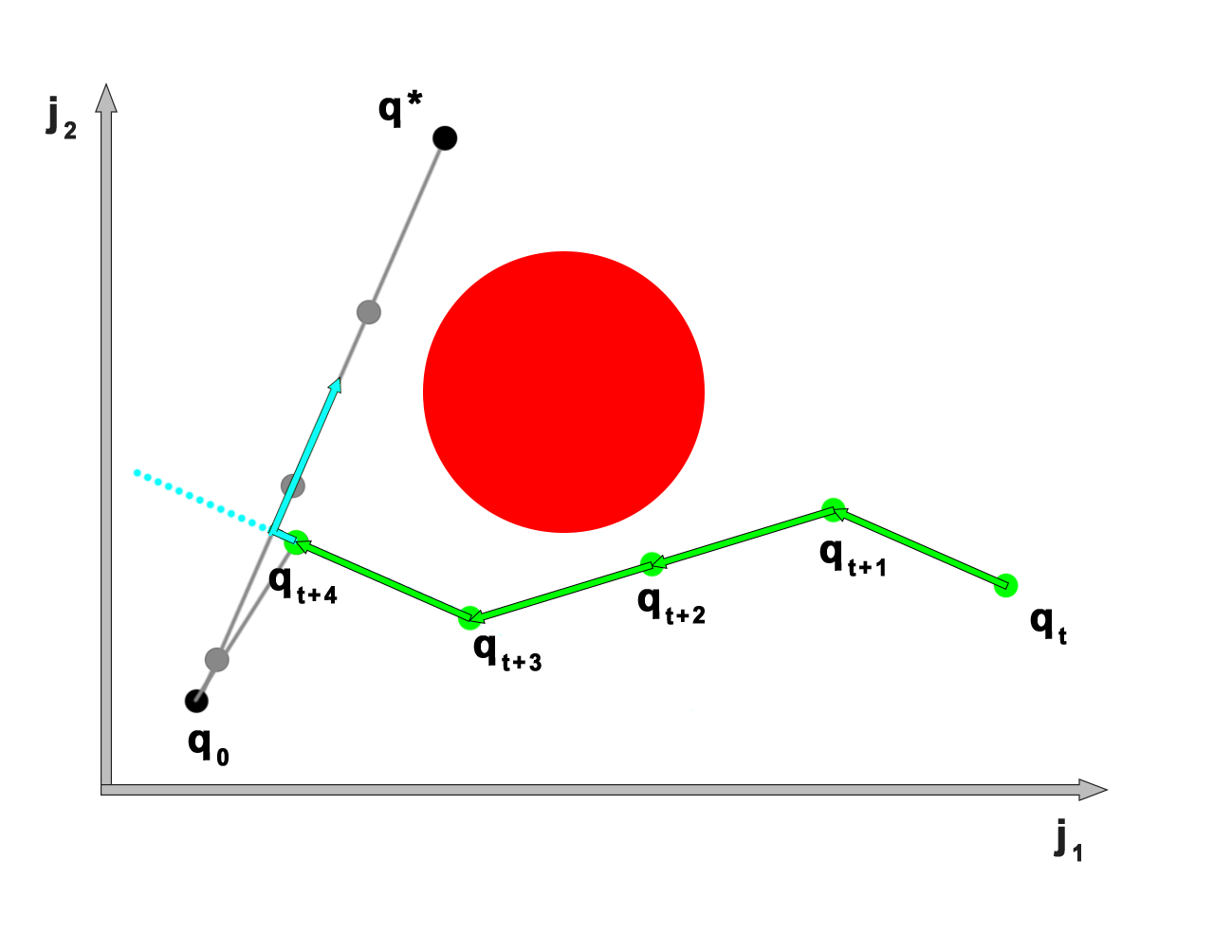}
		\vspace{-25pt}
		\subcaption{After four backsteps we can take the path to our target trajectory.}
	\end{subfigure}
	\hfill
	\begin{subfigure}[t]{0.23\textwidth}
		\centering
		\includegraphics[width=1.\textwidth]{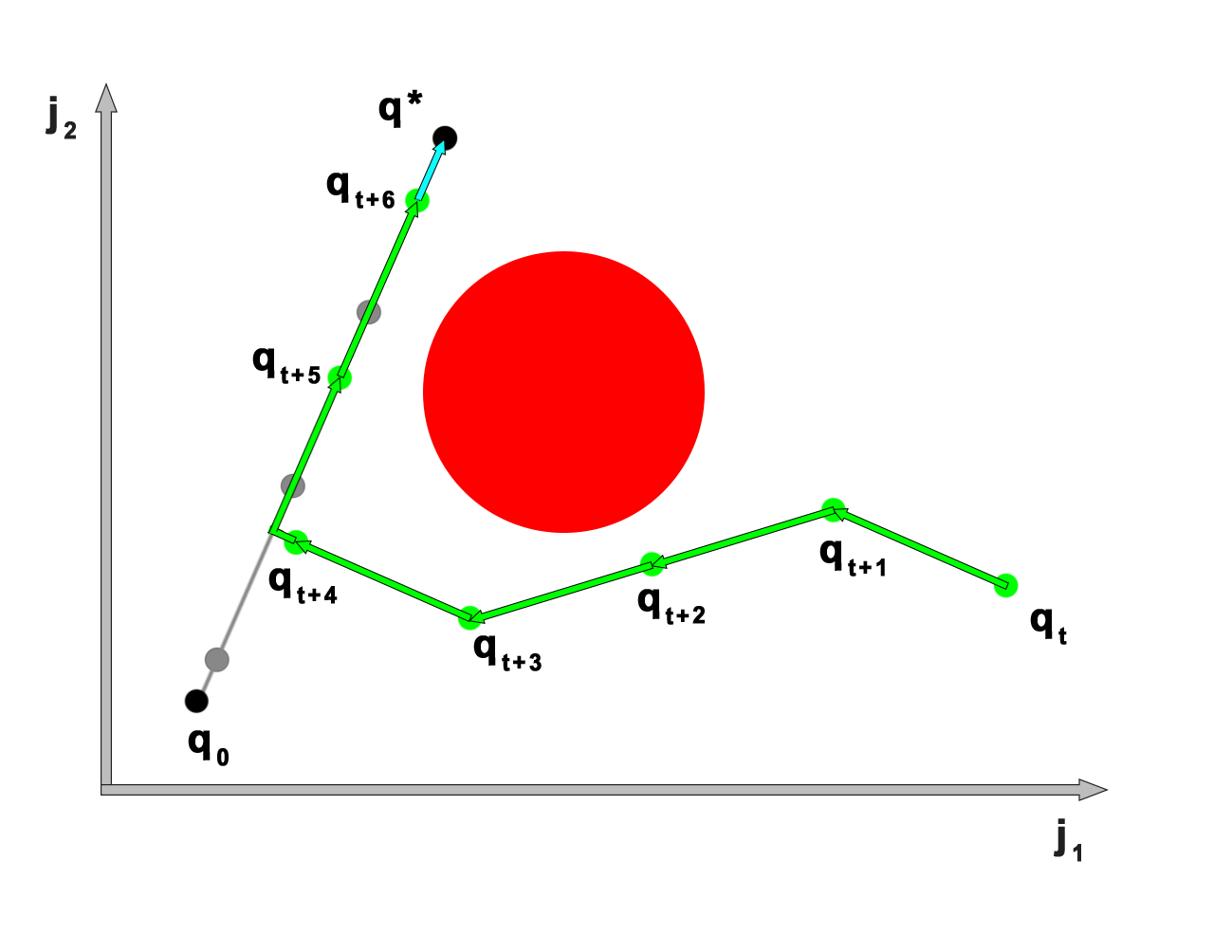}
		\vspace{-25pt}
		\subcaption{We follow the target trajectory towards $q^{\ast}$.}
	\end{subfigure}
	\caption{\centering Procedure of backstepping if collisions are encountered on the linear interpolation between $q_t$ and $q^{\ast}$.}
	\label{fig:backstepping}
	\vspace{-20pt}
\end{figure*}
\subsection{Grasp Configuration Observer}

We propose a dynamic grasp configuration observer to find a variety of $N$ grasp candidates for a cuboid-shaped object.
Since our tracking module outputs the pose of a cuboid-shaped bounding box around an object, this approach is also used to manipulate arbitrarily shaped objects.

Let $q\in\RRR^n$ be the robot configuration.
A grasp configuration candidate $g = (q, c, a)$ is defined by a joint configuration $q$, a cost $c\in\RRR$, and an age $a \in \NNN$.
Grasp candidates are determined by minimizing an inverse kinematics problem
\begin{align}
	q^* = \argmin_{q\in\mathbb{R}^n} \Vert q\Vert^2_W + \sum_{i=1}^M \lambda_i\left\Vert \phi_i(q)\right\Vert^2,
\end{align}
where $\phi_i:~\RRR^n \rightarrow \RRR^{d_i},~~y_i = \phi_i(q)$
are task maps that map a robot configuration to a $d_i$-dimensional space and $\lambda_i\in\RRR$ their weighting factors.
$W$ is a positive definite matrix providing regularization.

The following cost terms are considered during optimization.

\paragraph{Position}
The position cost is defined as
\begin{equation}
\phi_{pos}(q) =  p_o(q) - p_e(q),
\end{equation}
where $p_o: \RRR^n \rightarrow \RRR^3$ and $p_e: \RRR^n \rightarrow \RRR^3$ are functions mapping the configuration $q$ to the position of the target object and the end-effector, respectively.
\paragraph{Alignment}
Performing successful grasipng requires the end-effector to be aligned with the object.
We formulate the alignment cost as
\begin{align}
\phi_{align}(q) = &\left(1-\left(v_{o,x}(q)^Tv_{e,x}(q)\right)^2\right)\\ &\left(1-\left(v_{o,y}(q)^Tv_{e,x}(q)\right)^2\right) \\ &\left(1-\left(v_{o,z}(q)^Tv_{e,x}(q)\right)^2\right),
\end{align}
where $v_{o,x}: \RRR^n \rightarrow \RRR^3$ maps the configuration to the unit vector described by the $x$-axis of the target object coordinate system, analogously $v_{o,y}(q)$ and $v_{o,z}(q)$ for the $y$- and $z$-axis.
$v_{e, x}(q)$ maps to the unit vector describing $x$-axis of the end-effector coordinate system.
Minimizing this cost term aligns $v_{e,x}$ to one of the axes of the target object coordinate system.
\paragraph{Collision}
In order to successfully grasp the object, we need to avoid collisions with the object itself.
The collision cost term is defined as
\begin{equation}
\phi_{coll}(q) = (d(q) - m)[d(q) < m] 
\end{equation}
where $d$ maps $q$ to the pairwise distances of all shapes of the robot and the target object.
$m$ is an upper threshold on the minimum distance.
If the distance of two shapes is less than $m$ this cost term becomes active.
\paragraph{Joint Limits \& Homing}
We further add a cost term $\phi(q)_{limit}$ increasing when joint limits of the robot are violated and a homing cost term $\phi(q)_{home}$.

$q^{\ast}$ is the configuration of a feasible grasp.
However, computing $N$ grasp candidates starting at the same configuration $q_0$ results in all candidates defining the same configuration $q^{\ast}$.
This is especially relevant for the alignment term $\phi_{align}$, which has multiple local minima.
Instead of explicitly enumerating possible different alignments as in \cite{19-driess-RSSws, driess2020deepRSS}, we diversify grasp candidate computation by introducing a random alignment term, effectively randomizing the approach axis of a grasp candidate. 
That is, at initialization we compute a random orthonomal basis
\begin{equation}
R =  \begin{pmatrix}v_1 & v_2 & v_3\end{pmatrix} \in \RRR^{3 \times 3}.
\end{equation}
We use $R$ to introduce an additional cost term
\begin{align*}
	\begin{split}
	\phi_{rand}(q) = 
	\begin{pmatrix}
	1-\left(v_1(q)^Tv_{x,e}(q)\right)\left(v_1(q)^Tv_{x,e}(q)\right) \\
	1-\left(v_2(q)^Tv_{y,e}(q)\right)\left(v_2(q)^Tv_{y,e}(q)\right) \\
	1-\left(v_3(q)^Tv_{z,e}(q)\right)\left(v_3(q)^Tv_{z,e}(q)\right)
	\end{pmatrix}.
	\end{split}
\end{align*}
We further introduce two additional weighting functions
\begin{align}
f_{inc}(a) &= \frac{1}{1+\exp(\frac{\alpha}{2}-a)}\\
f_{dec}(a) &= \frac{1}{1+\exp(a-\frac{\alpha}{2})}
\end{align}
depending on the age $a$ of a grasp candidate. $\alpha$ is a constant threshold.
During optimization we weight $\phi_{pos}(q),~\phi_{align}(q),~\phi_{coll}(q)$ with $f_{inc}(a)$ and $\phi_{rand}(q)$ with $f_{dec}(a)$ ensuring that only the approach axis is randomized and no force balance arises between $\phi_{align}(q)$ and $\phi_{rand}(q)$.

\subsection{Grasp Candidate Ranking}

Choosing the best grasp from $N$ grasp candidates is non-trivial.
The essential and hard requirement for a good grasp candidate is whether the grasp configuration ends in an end-effector position inside the target cuboid.
We compute this using the end-effector position relative to the cuboids coordinate system and the shape of the cuboid.
We assume that success of grasping increases with decreasing distance to the center of the cuboid.
If two candidates are both inside the cuboid and also share the same distance to the center, we choose the one with lower homing costs. 

\subsection{Updating Grasp Candidates}

Since an online reactive grasping approach needs to take the changing environment into account, we continously update the grasp candidates.
At initialization all grasp candidates have the same cost vector $c$.
In each iteration the age $a$ of each grasp is incremented and the cost vector of each grasp is evaluated using the current configuration $q_t$.

\subsection{Trajectory Planning}

Our goal is to follow a trajectory starting in the current configuration $q_{t}$ and ending in the configuration $q^{\ast}$ of the best grasp candidate $q^{\ast}$.
Due to the possibilty of a changing environment, and more specifically the change of the target object pose, we cannot plan a full trajectory in advance.
Movements of the target object can lead to collisions during a grasp approach following a pre-planned trajectory.
We mitigate this problem by using a technique we call \textit{backstepping}.

Our goal is to follow a target trajectory $T$ defined by the linear interpolation with stepsize $\tau$ between the homing configuration $q_0$ and the configuration $q^{\ast}$ of the best grasp candidate.
At each step we find the closest configuration $q_s$ on the target trajectory $T$ to our current configuration $q_t$.
If a step of stepsize $\tau$ ends up in a collision with an object, we take a step towards $q_0$.
The procedure of backstepping is depicted in figure~\ref{fig:backstepping}.
\section{Experiments}\label{sec:experiments}
This section describes experiments conducted on object tracking using \methodname{} in simulation and experiments using the complete system for reactive grasping on a real robot.

\subsection{Object Tracking}
For quantitive comparison to an existing method we conduct experiments in simulation.
We render a camera view on the scene which is located at the same pose as our camera on the real world setup.
We compare \textcolor{flatgreen}{\methodname{}} to a particle-filter based object tracking method from the PCL library~\cite{Tracking3DObjects} (named \textcolor{flatpurple}{particle} in the plots).
Errors are reported in terms of position and rotation.
Rotational errors are reported in terms of the geodesic loss.

\begin{figure}[!t]
  \begin{subfigure}[b]{0.238\textwidth}
    \centering
    \includegraphics[width=\textwidth]{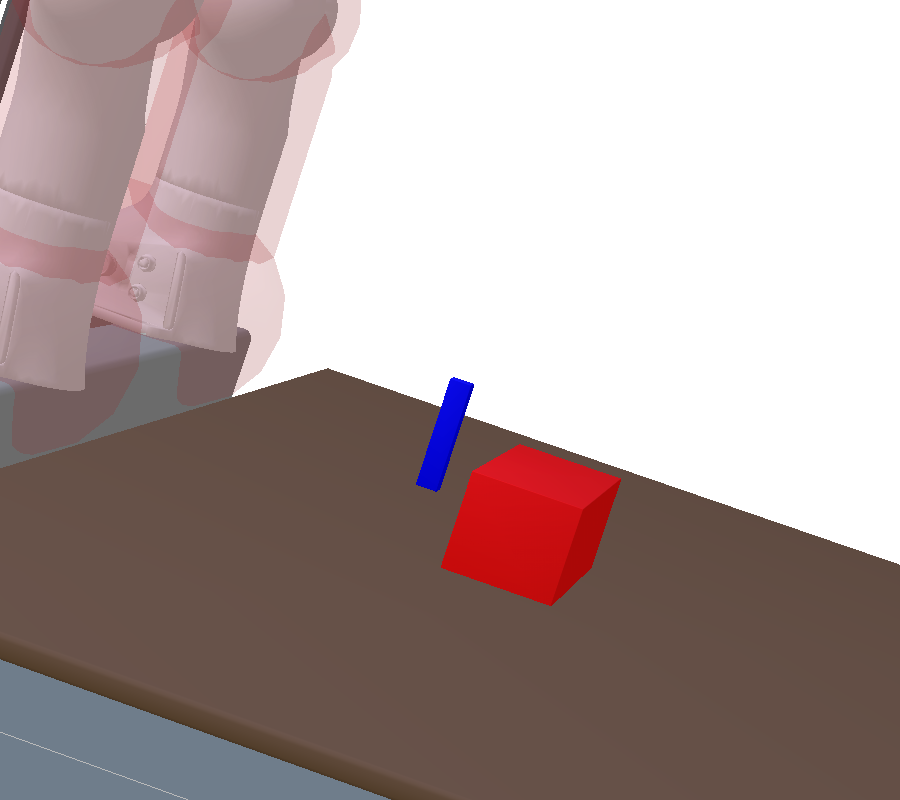}
    \caption[]%
    {A small rectangle occludes the target object.}
    \label{fig:experiments:occlusion:a}
  \end{subfigure}
  \hfill
  \begin{subfigure}[b]{0.238\textwidth}
    \centering
    \includegraphics[width=\textwidth]{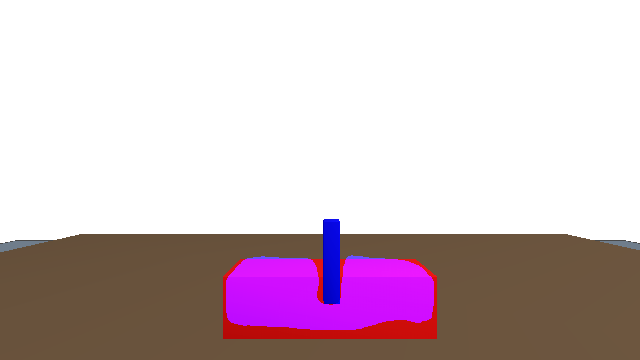}
    \caption[]%
    {Camera view on the scene. The object, which occludes the target object, is not contained in the segmentation mask.}
    \label{fig:experiments:occlusion:b}
  \end{subfigure}
  \vskip\baselineskip
    \vspace{-10pt}
    \centering
  \begin{subfigure}[b]{.5\textwidth}
    \centering
    \includegraphics[trim=0 0 0 50,clip,width=\linewidth]{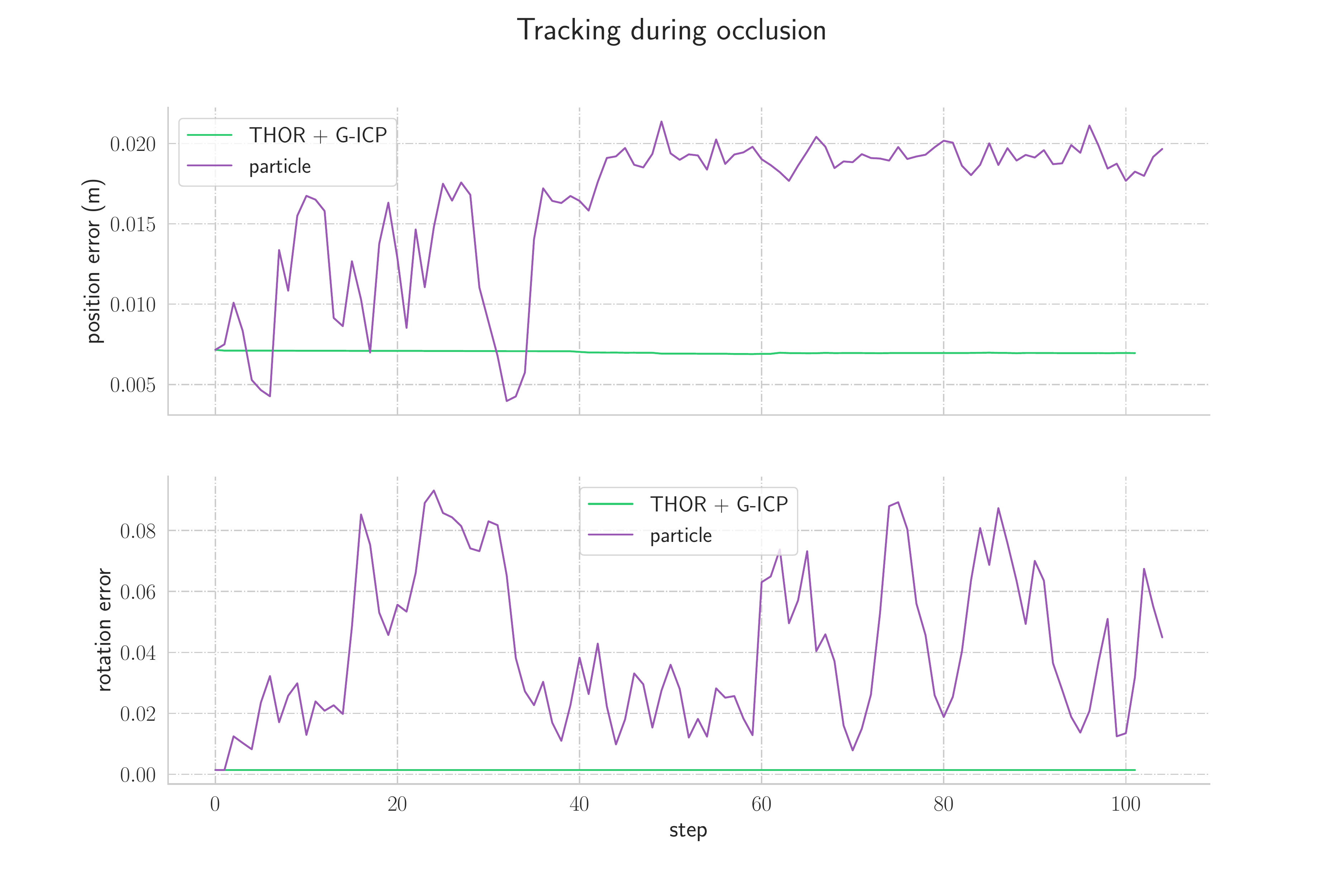}
    \vspace{-23pt}
    \caption[Object tracking under occlusion: results]{Results of object tracking during occlusion. \methodname{} remains stable during the complete tracking sequence.}
    \label{fig:experiments:occlusion:results}
  \end{subfigure}
  \vspace{-10pt}
  \caption[Object tracking under occlusion.]{Object tracking during occlusion.}
  \label{fig:experiments:occlusion}
  \vspace{-20pt}
\end{figure}

\paragraph{Occlusion}
Occlusions occur frequently during robotic manipulation.
When grasping an object with a parallel jaw gripper, parts of the object are already occluded by the gripper fingers.
In this experiment a small box moving into the scene from the top occludes the target object, as seen in figure~\ref{fig:experiments:occlusion:a}.
Results, seen in figure~\ref{fig:experiments:occlusion:results}, show that occlusion has almost no impact on the performance of \methodname{}.
The accuracy of the particle tracker, however, is reduced proportional to the amount of occlusion.
The stability of \methodname{} relies on the fact that SiamMask is able to accurately segment the tracked object from the occluding object.
As seen in figure~\ref{fig:experiments:occlusion:b}, the segmentation mask of SiamMask contains parts of the target object only. 
G-ICP is able to fit the template pointcloud to the observations seamlessly, since the observation pointcloud omits the occluded part.

\paragraph{Manipulation}

\begin{figure}[!t]
  \begin{subfigure}[b]{0.238\textwidth}
    \centering
    \includegraphics[width=\textwidth]{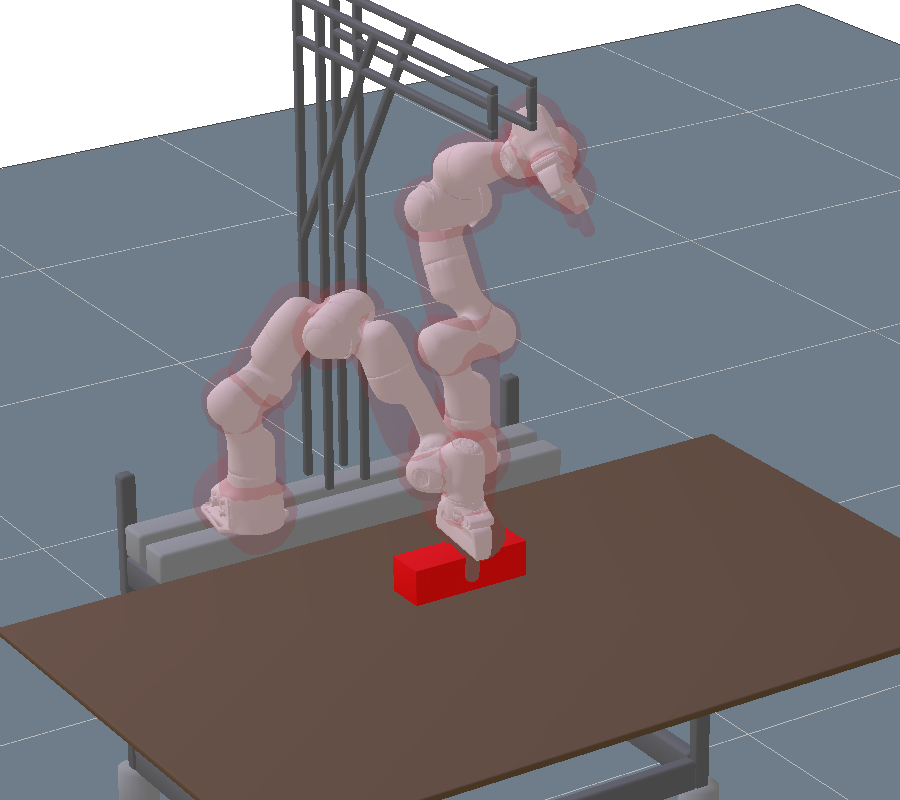}
 \vspace{-15pt}
    \caption[]%
    {{}}
    \label{fig:experiments:manipulation:a}
  \end{subfigure}
  \hfill
  \begin{subfigure}[b]{0.238\textwidth}
    \centering
    \includegraphics[width=\textwidth]{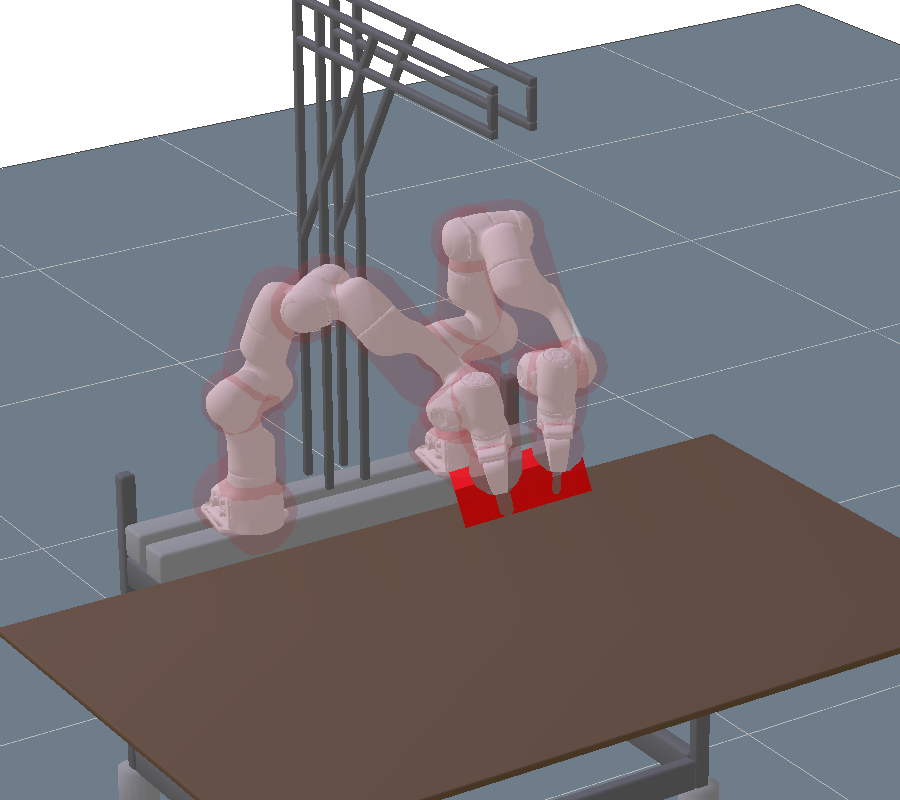}
 \vspace{-15pt}
    \caption[]%
    {{}}
    \label{fig:experiments:manipulation:b}
  \end{subfigure}
  \vskip\baselineskip
  \begin{subfigure}[b]{0.238\textwidth}
    \centering
    \includegraphics[width=\textwidth]{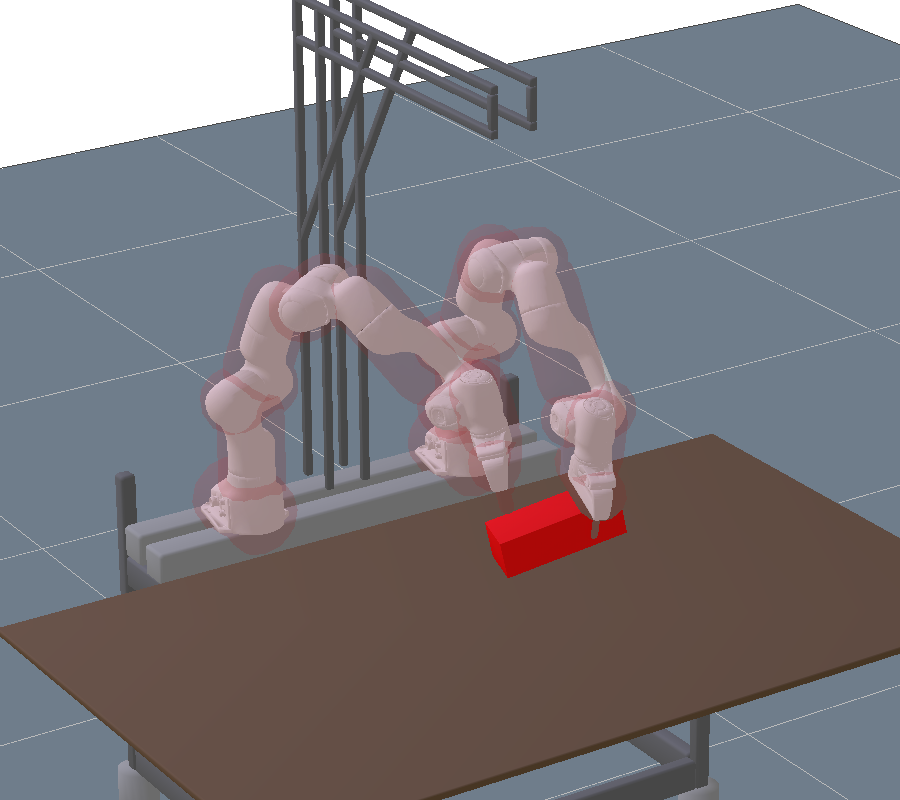}
 \vspace{-15pt}
    \caption[]%
    {{}}
    \label{fig:experiments:manipulation:c}
  \end{subfigure}
  \hfill
  \begin{subfigure}[b]{0.238\textwidth}
    \centering
    \includegraphics[width=\textwidth]{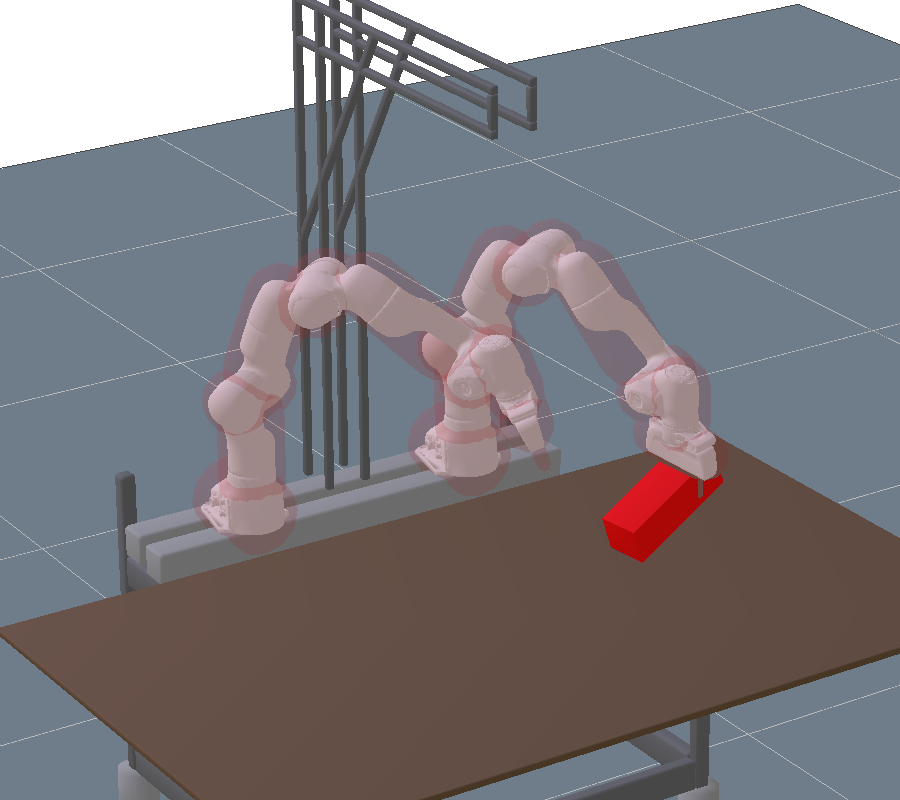}
 \vspace{-15pt}
    \caption[]%
    {{}}
    \label{fig:experiments:manipulation:d}
  \end{subfigure}
  \vskip\baselineskip
\begin{subfigure}[b]{.5\textwidth}
  \centering
 \vspace{-10pt}
  \includegraphics[width=\textwidth]{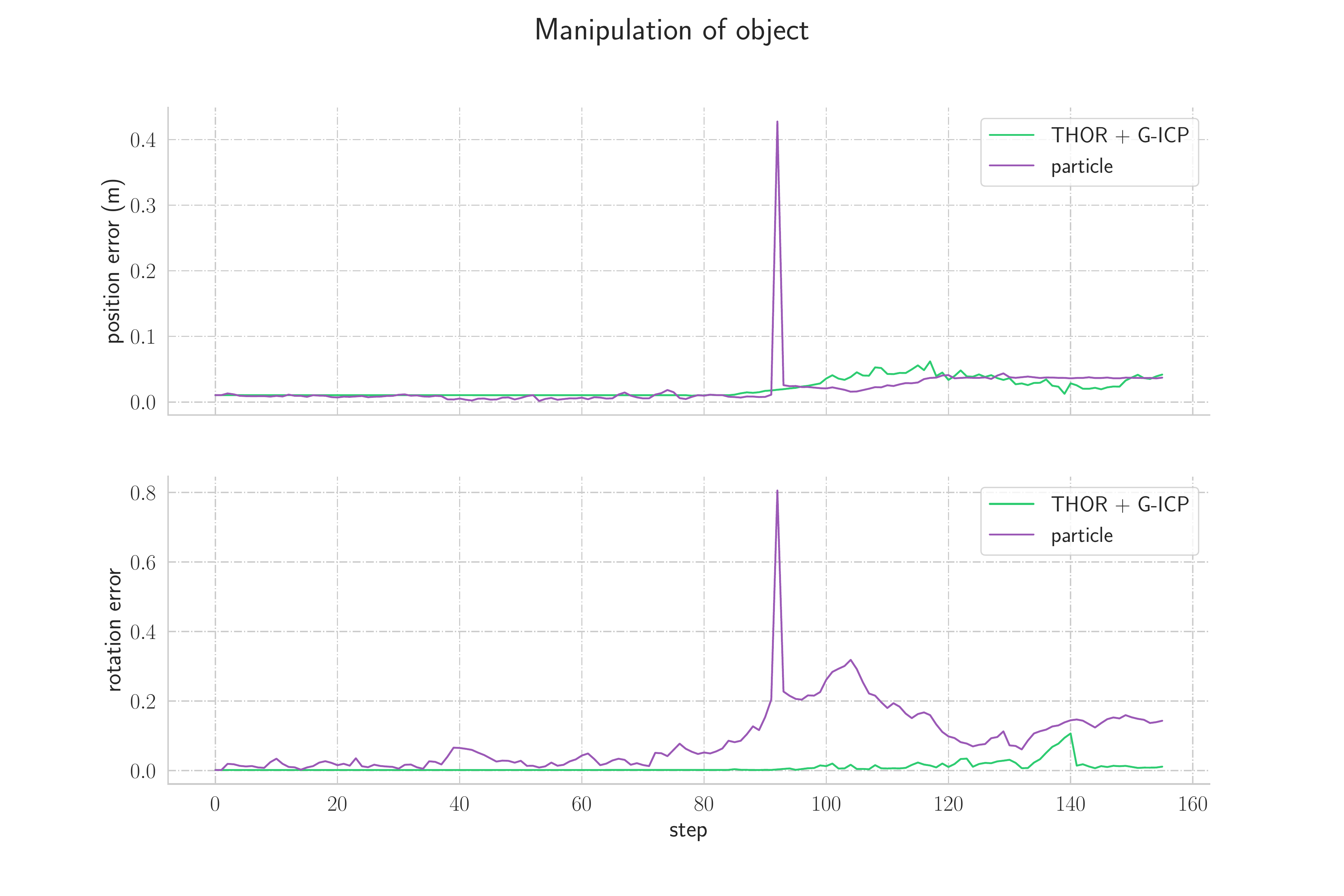}
 \vspace{-20pt}
  \caption[Results of tracking during manipulation]{Results of tracking during manipulation.}
  \label{fig:experiments:manipulation:results}
\end{subfigure}
  \caption[Object tracking during manipulation.]{Object tracking during manipulation. In a) the first robot picks up the target object and hands it over to the second robot in b). The second robot moves the object to an intermediate position in c) and finally moves to the goal position d) while keeping the object in hand.}
  \label{fig:experiments:manipulation}
  \vspace{-20pt}
\end{figure}

Supporting robotic manipulation is the main purpose of our object tracking method and therefore an important experiment.
It incorporates many different possibilites which can occur during object tracking, e.g. translational and orientational movements as well as partial occlusion due to the gripper fingers or the second robot.
Figure~\ref{fig:experiments:manipulation} shows the manipulation process and results of object tracking during manipulation.
Results of object tracking are depicted in figure~\ref{fig:experiments:manipulation:results}.
These show that both methods are able to track the target object during manipulation.
However, the particle filter tracker loses the object once completely.
This corresponds to the moment of the handover (figure~\ref{fig:experiments:manipulation:b}) and is due to the occlusion both gripper fingers generate on the object.
Tracking with \methodname{} is stable during the whole manipulation process and also generates almost no rotational error.

\subsection{Reactive Grasping}
We perform reactive grasping with a Franka Emika Panda Robot.
Objects on the table are detected from a top down view using a \textit{ASUS Xtion PRO LIVE RGBD} camera.
Once an object on the table is detected, \methodname{} is initialized with an \textit{Intel RealSense D415} camera from a side view.
These experiments are also shown in our accompanying video.

\paragraph{Unknown Objects}
Experiments on grasping of unknown objects are conducted in this section. 
Figure~\ref{fig:objectstograsp} shows the objects which are manipulated in our experiments.
\methodname{} is able to track each of these objects during the complete grasping process.
Further, the approach to reactive grasping presented in this work is able to grasp each object even under disturbances.
The pringles can is moved by a human during the manipulation process.
However, our approach is able to grasp the object successfully.
Notably, the gripper finger occludes the measuring tape by more than half of its size in the camera view.
The lashing strap is surrounded by a plastic firm, exposing heavy reflections.
Although reflections impose difficulties on classical computer vision techniques, \methodname{} is able to track the lashing strap effortlessly.

\paragraph{Collaborative Manipulation}
Experiments on grasping under heavy disturbances are conducted.
During theses experiment the robot and the target object are moved and re-oriented by a human during grasp approaches.
The robot is also prevented from grasping the object and moved to a position beside the object, likely leading to collisions with the object. 
Target objects are successfully tracked throughout the process and the path planning component is able to recover from each state, leading to a successful grasp in each experiment.
\section{Conclusion}\label{sec:conclusion}
We present a system to track and dynamically manipulate unknown objects.
Our experiments show that our approach to reactive grasping is successful even under occlusion of the target object and under perturbation of either the robot or the object.
\bibliographystyle{IEEEtran}
\bibliography{IEEEabrv, root}




\end{document}